\documentclass[conference]{IEEEtran}
\IEEEoverridecommandlockouts
\usepackage{cite}
\usepackage{amsmath,amssymb,amsfonts}
\usepackage{algorithmic}
\usepackage{graphicx}
\usepackage{textcomp}
\usepackage{enumitem}
\usepackage{multirow}
\usepackage{booktabs} 
\usepackage{xcolor}
\def\BibTeX{{\rm B\kern-.05em{\sc i\kern-.025em b}\kern-.08em
    T\kern-.1667em\lower.7ex\hbox{E}\kern-.125emX}}
\begin{document}

\title{Leveraging Large Language Models for Patient Engagement: The Power of Conversational AI in Digital Health}

\author{\IEEEauthorblockN{Bo Wen}
\IEEEauthorblockA{\textit{Digital Health} \\
\textit{IBM T. J. Watson Research Center}\\
Yorktown Heights, NY, USA \\
bwen@us.ibm.com}
\and
\IEEEauthorblockN{Raquel Norel}
\IEEEauthorblockA{\textit{Digital Health} \\
\textit{IBM T. J. Watson Research Center}\\
Yorktown Heights, NY, USA \\
rnorel@us.ibm.com}
\and
\IEEEauthorblockN{Julia Liu}
\IEEEauthorblockA{\textit{Gastroenterology} \\
\textit{Morehouse School of Medicine}\\
Atlanta, GA, USA \\
jjliu@msm.edu}
\and
\IEEEauthorblockN{Thaddeus Stappenbeck}
\IEEEauthorblockA{\textit{Inflammation and Immunity} \\
\textit{Cleveland Clinic Foundation}\\
Cleveland, OH, USA \\
stappet@ccf.org}
\and
\IEEEauthorblockN{Farhana Zulkernine}
\IEEEauthorblockA{\textit{School of Computing} \\
\textit{Queen's University}\\
Kingston, Ontario, Canada \\
farhana.zulkernine@queensu.ca}
\and
\IEEEauthorblockN{Huamin Chen}
\IEEEauthorblockA{\textit{Office of CTO} \\
\textit{Red Hat}\\
Boston, MA, USA \\
hchen@redhat.com}
}

\maketitle

\begin{abstract}
The rapid advancements in large language models (LLMs) have opened up new opportunities for transforming patient engagement in healthcare through conversational AI. This paper presents an overview of the current landscape of LLMs in healthcare, specifically focusing on their applications in analyzing and generating conversations for improved patient engagement. We showcase the power of LLMs in handling unstructured conversational data through four case studies: (1) analyzing mental health discussions on Reddit, (2) developing a personalized chatbot for cognitive engagement in seniors, (3) summarizing medical conversation datasets, and (4) designing an AI-powered patient engagement system. These case studies demonstrate how LLMs can effectively extract insights and summarizations from unstructured dialogues and engage patients in guided, goal-oriented conversations. Leveraging LLMs for conversational analysis and generation opens new doors for many patient-centered outcomes research opportunities. However, integrating LLMs into healthcare raises important ethical considerations regarding data privacy, bias, transparency, and regulatory compliance. We discuss best practices and guidelines for the responsible development and deployment of LLMs in healthcare settings. Realizing the full potential of LLMs in digital health will require close collaboration between the AI and healthcare professionals communities to address technical challenges and ensure these powerful tools' safety, efficacy, and equity.
\end{abstract}


\section{Introduction}
The field of digital health is undergoing a rapid transformation fueled by the convergence of advanced technologies and the increasing availability of health data. Large language models (LLMs), deep learning models trained on vast amounts of text data, have emerged as a particularly promising tool for unlocking insights and automating tasks across various healthcare domains. With their ability to understand, generate, and reason with natural language, LLMs are enabling new applications and reshaping existing practices in healthcare.

This paper aims to provide a comprehensive overview of LLMs' current state and future directions in digital health. We begin by discussing the recent breakthroughs in LLM architectures and training techniques that have led to significant improvements in performance and capabilities. Next, we present four case studies showcasing the diverse applications of LLMs in healthcare, spanning conversation data mining, medical conversation analysis, and patient engagement. Through these examples, we demonstrate the potential of LLMs to address real-world challenges and improve patient outcomes.

However, integrating LLMs into healthcare also raises important ethical considerations and challenges. We explore issues related to data privacy, bias and fairness, transparency and interpretability, and regulatory compliance. We provide guidelines for responsibly developing and deploying LLMs in healthcare settings based on the latest research and best practices.

Considering future possibilities, we identify emerging trends and opportunities for LLMs in digital health. These include integrating LLMs with other technologies, such as the Internet of Medical Things (IoMT) \cite{Mehmood2020ManagingDD} and blockchain \cite{he2024large, luo2023bc4llm}, developing specialized medical LLMs \cite{chan2023assessing}, and applying LLMs to new areas such as personalized medicine and drug discovery. We also highlight the importance of interdisciplinary collaboration and the need for standardized evaluation frameworks to assess the quality and impact of LLM-based interventions.

\section{Recent Advancements in Large Language Models}
The field of large language models has witnessed significant breakthroughs in the past year, with the release of more powerful and capable models like GPT-4 \cite{openai2023gpt4}, LLaMa \cite{touvron2023llama}, and PaLM \cite{chowdhery2022palm}. These models push the boundaries of what is possible with natural language processing (NLP), thanks to architectural innovations, training techniques, and the scale of data and computing power used.

One notable advancement is the ability of LLMs like GPT-4 to handle multimodal data, such as images and text, enabling new applications that combine visual and linguistic understanding. Another critical development is the rise of open-source LLMs like the LLaMa family \cite{touvron2023llama, touvron2023llama2}, Falcon \cite{almazrouei2023falcon}, and the Mistral family \cite{jiang2023mistral, jiang2024mixtral}, which are helping to democratize access to these powerful tools and spurring further innovation.

Instruction tuning \cite{wei2022finetuned, zhang2024instruction} and reinforcement learning from human feedback (RLHF)\cite{instructGPT, christiano2017deep} have also emerged as standard techniques for aligning LLMs with desired behaviors and outputs. Researchers can create more controllable and helpful language models by explicitly training LLMs to follow instructions and optimizing them based on human preferences.

Recent advancements in LLMs have enabled breakthroughs in multimodal learning, particularly in text-to-image and text-to-video generation \cite{videoworldsimulators2024}. These developments build upon progress in both autoregressive models \cite{ramesh2021zero, yu2022scaling} and diffusion models \cite{ramesh2022hierarchical, rombach2022high, nichol2022glide, saharia2022photorealistic}, which leverage the power of large-scale pre-trained language and vision models to generate high-quality visual content from textual descriptions. These models' scale and architectural improvements have significantly enhanced the quality and coherence of the generated images and videos and their ability to capture complex language semantics \cite{yu2022scaling, saharia2022photorealistic}.

These advances in multimodal learning demonstrate the potential for LLMs to enable new forms of interactive communication \cite{yang2023dawn,wu2023visual}. In digital health, multimodal LLMs could be harnessed to enhance patient engagement and education by generating personalized visual content to accompany text-based interventions. For example, in cognitive assessment tools like the Cookie Theft picture description task \cite{goodglass2001bdae}, LLMs could be used to generate customized visual cues tailored to a patient's cultural background, interests, or cognitive abilities, potentially improving the sensitivity and specificity of the test. Similarly, multimodal LLMs could create interactive visual aids to support patient education materials, medical instructions, or treatment adherence reminders, making the information more engaging and easier to understand.

As LLMs continue to evolve and incorporate more modalities \cite{girdhar2023imagebind, zhang2023metatransformer}, they may open up further possibilities for integrating physiology signals, medical images, wearable and IoMT signals into patient-provider communication and personalizing healthcare interventions.

Despite the immense potential, there are still technical challenges and risks to address as LLMs are applied more widely in medicine \cite{nori2023capabilities, 10261199}. For example, improving inference speed and efficiency to enable real-time use through techniques like model compression, quantization, pruning, and system-level optimizations; ensuring the reliability, factuality and lack of bias in LLM outputs, especially for high-stakes medical decisions; protecting patient privacy and securing sensitive medical data used to train LLMs; validating LLM performance through extensive clinical testing and benchmarking; and establishing regulatory frameworks and best practices for the development and deployment of LLMs in healthcare settings, are some of the open research challenges.

The ongoing research and engineering efforts to make LLMs faster, lighter, and safer, are poised to transform many aspects of medical practice and enable a new era of AI-assisted healthcare. 
However, realizing the full potential of multimodal LLMs in healthcare will require careful consideration of ethical, privacy, and safety implications and close collaboration between AI researchers and healthcare stakeholders to ensure the responsible development and deployment of these powerful tools.

\section{Case Studies: Applications of LLMs in Patient Engagement}
This section presents four case studies from the IBM Digital Health department. They demonstrate the power of LLMs in analyzing and generating conversations for improved patient engagement. These studies highlight how LLMs can effectively handle unstructured conversational data, a task that was traditionally challenging before the advent of these powerful models.

\subsection{Case Study 1: Analyzing Mental Health Discussions on Reddit}
This study \cite{Bauer2024} applied LLMs to investigate linguistic patterns in mental health discussions on Reddit, focusing on the ``Suicide Watch'' subreddit, which provides a digital space for individuals struggling with suicidal thoughts to share their thoughts and ask for advice. By analyzing 2.9 million posts across 30 subreddits from October to December 2022 and the corresponding period in 2010, the study aimed to identify themes consistent with known risk factors for suicidal ideation.

The research leverages LLMs and natural language processing (NLP) to analyze online discussions about suicidality on Reddit, showcasing a pivotal application of conversational AI in healthcare. By employing sophisticated AI methodologies to interpret the nuanced language used in mental health forums, the study highlights how conversational AI can be instrumental in identifying and understanding mental health issues through user-generated content. This not only facilitates real-time monitoring and support but also enhances the responsiveness of digital health interventions.

The findings underscore the potential of conversational AI to provide valuable insights into the emotional and psychological states of individuals, which is crucial for developing targeted interventions and improving patient engagement in digital healthcare platforms. Through such advanced analysis, conversational AI can evolve to become more empathetic and effective, aligning closely with the needs and conditions of those it aims to serve. This reinforces its role in proactive mental health management and support.

\begin{figure}[!ht]
\centering
\includegraphics[width=0.7\linewidth]{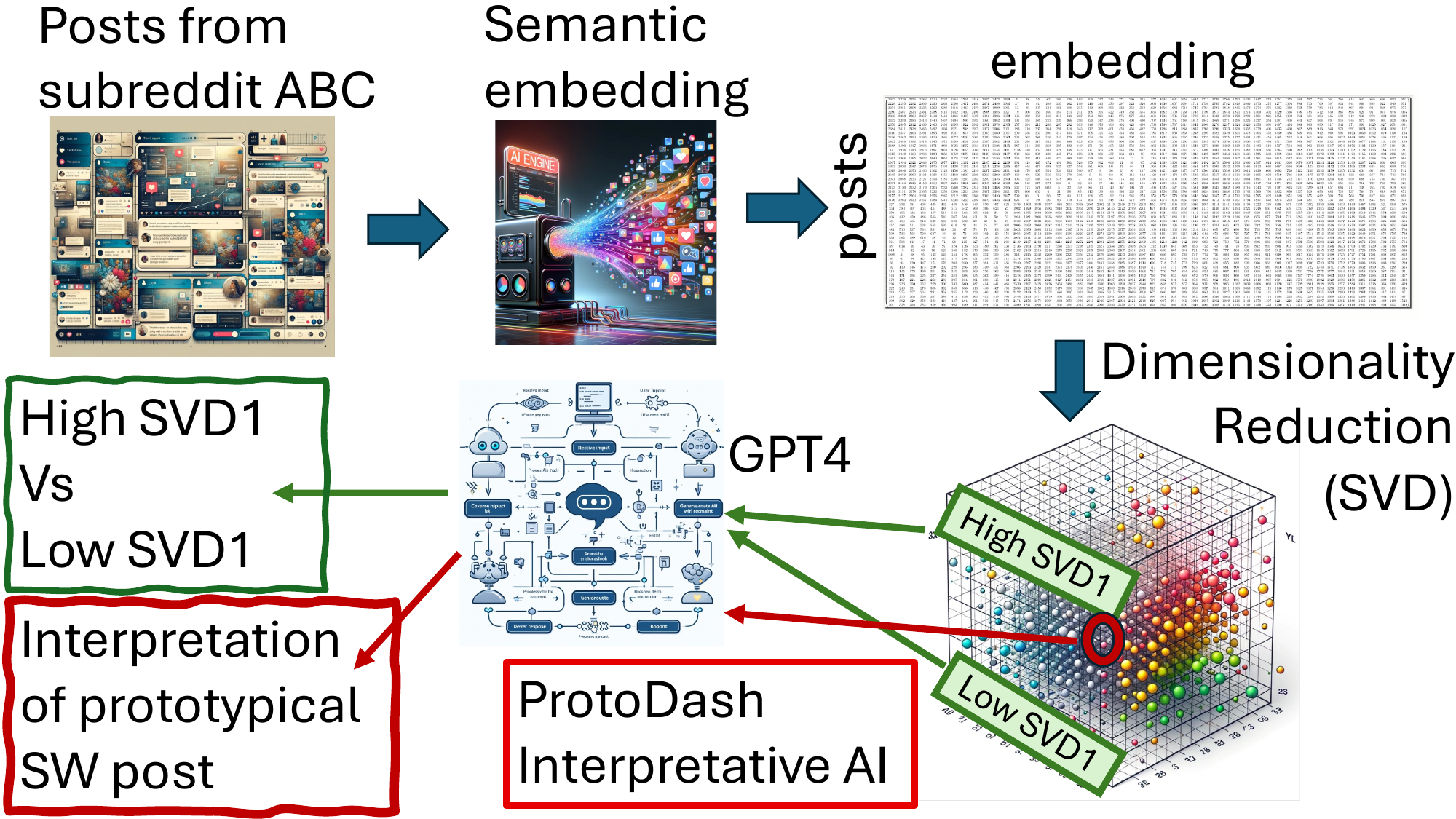}
\caption{Overview of study methodology. The sub-images were created with the help of DALL$\cdot$E 3.}
\label{fig:SW_DALLE}
\end{figure}

The study involved using LLMs to generate numerical representations (embeddings) of the posts, capturing semantic similarities between the text content. Dimensionality reduction techniques were then applied to these embeddings to identify latent linguistic dimensions and visualize the relationships between different subreddits. For an overview of the methodology, see Figure \ref{fig:SW_DALLE}. This approach highlights the potential of conversational AI to analyze unstructured data and extract clinically meaningful insights, which can inform AI-driven mental health interventions.

The analysis revealed several key findings. First, posts in the Suicide Watch subreddit were characterized by linguistic themes consistent with known risk factors for suicidal ideation, including perceived burdensomeness, hopelessness, disconnection, and trauma (see Table 2 from \cite{Bauer2024}). These themes align with the interpersonal theory of suicide \cite{VanOrden2008, van2010interpersonal} and other ``ideation-to-action'' frameworks \cite{klonsky2015three, oconnor2011towards}, providing empirical support for these theories based on naturalistic social media data.

Furthermore, the study identified distinct linguistic dimensions correlating with well-being, the pursuit of support, and the severity of distress. 
This groundbreaking approach not only corroborates existing theories of suicide and mental health disorders but also paves the way for a deeper, data-driven understanding of the intricate web of factors contributing to suicidality, showcasing the transformative potential of explainable AI in mental health research.

Second, the study identified three latent linguistic dimensions across the mental health subreddits: emotional well-being (ranging from despair to resilience), support seeking (from self-directed coping to actively requesting help), and symptom severity (from mild distress to acute crisis). Subreddits focused on different mental health conditions mapped onto these dimensions in distinct clusters (Figure 5-6 from \cite{Bauer2024}), suggesting a shared linguistic structure underlies discussions of various disorders. Notably, the Suicide Watch subreddit scored high on the negative pole of well-being, the support-seeking pole, and the severe distress pole, reflecting the high-risk state of its user base.

Third, the clustering of subreddits along these dimensions corresponded to the proposed structure of psychopathology in the Hierarchical Taxonomy of Psychopathology (HiTOP) \cite{kotov2017hierarchical}. Subreddits related to internalizing disorders (e.g., anxiety, depression, PTSD), thought disorders (e.g., schizophrenia, psychosis), and externalizing disorders (e.g., addiction, BPD) formed distinct clusters (Figure 7 from \cite{Bauer2024}), providing data-driven validation of this hierarchical, dimensional model of mental disorders based on linguistic patterns in social media.

The study demonstrates the potential of LLMs to extract clinically meaningful insights from unstructured online discussions, which may help identify at-risk individuals, inform assessment and intervention strategies, and monitor population-level mental health trends. However, this study is a proof-of-concept and it does have several limitations: The data only spans a three-month period, which may not capture longer-term patterns or seasonal variations. The study is limited to Reddit and may not generalize to other social media platforms or offline conversations. The LLMs were not specifically fine-tuned for mental health, so their ability to capture nuanced clinical concepts is unknown. Finally, the study did not have access to users' actual mental health histories or outcomes, so the findings are correlational and cannot prove the predictive validity of the identified linguistic markers.

Future work should aim to validate the linguistic patterns against ground-truth clinical data, examine their stability over time and across platforms, and assess their utility for guiding interventions and predicting individual-level clinical trajectories. Integrating multimodal data (e.g., images, metadata) and developing mental health-specific LLMs may further improve the precision and generalizability of the approach.

\subsection{Case Study 2: Personalized Chatbot for Cognitive Engagement in Seniors}
This ongoing study \cite{zhou2024bookclub} aims to investigate the potential of LLM-powered chatbots to promote reading engagement and prevent cognitive decline in older adults. The project involves developing a conversational AI agent that engages seniors in discussions about their reading interests and progress, with the hypothesis that regular, stimulating reading may help maintain cognitive function. The study is being conducted in collaboration with the Laboratory for Brain and Cognitive Health Technology at McLean Hospital and Harvard Medical School, and targets participants aged 70 and above residing in assisted living communities.

The participants are divided into a control group and an intervention group, with the chatbot serving as an assistive tool to encourage the intervention group to read more regularly and frequently than the control group and keep engaged with the material being read. To ensure a meaningful difference in reading behavior between the two groups, which is crucial for the accuracy of the study results, the chatbot is designed to make the reading experience more engaging, enjoyable, and rewarding for the intervention group participants. By providing a personalized, interactive platform for discussing books and exploring reading interests, the chatbot aims to replicate the stimulating  experience of a book club, which would be prohibitively expensive and logistically challenging to implement with human moderators in the context of a large-scale, longitudinal study. The chatbot's ability to tailor its conversational style, content recommendations, and discussion prompts to each user's unique preferences and needs is expected to enhance motivation, engagement, and adherence to the reading intervention throughout the study period. While the chatbot technology has potential applications for promoting reading habits in various age groups, its specific purpose in this study is to create a controlled environment where the effects of increased reading on cognitive health can be rigorously investigated in an older adult population. The use of an AI-powered assistant not only makes this type of study design feasible but also opens up new possibilities for personalized, scalable interventions to support healthy aging.

\begin{figure}[htbp]
    \centering
    \includegraphics[width=1\linewidth]{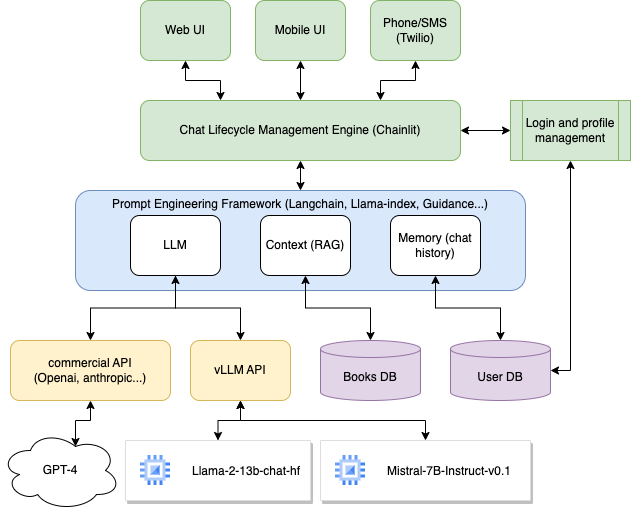}
    \caption{Architecture of the bookclub host chatbot. For more technical details, please visit the open-source repo: https://github.com/Twilight-Tales/Twilight-Chat and the technical report \cite{zhou2024bookclub}}
    \label{fig:bookclub-bot-arch}
\end{figure}

The chatbot, see figure \ref{fig:bookclub-bot-arch}, is built using LLMs and prompt engineering techniques to generate open-ended questions based on each user's reading profile and conversation history. The system maintains user profiles that track individuals' reading progress and key discussion points, allowing the chatbot to tailor its responses to each user's interests. For example, if a user expressed interest in a particular character or theme in a previous conversation, the chatbot may ask follow-up questions or draw connections to those elements in subsequent discussions.

To ensure accessibility and ease of use for the elderly population, the chatbot provides both voice (via phone calls) and text (via mobile webapp or SMS) interfaces. Preliminary user testing has yielded promising results and insights. Participants reported finding the conversations engaging and mentally stimulating, with some noting that the chatbot prompted them to think more deeply about their reading and make connections they might not have otherwise. Several users appreciated the convenience and flexibility of being able to chat with the bot at any time, without the need to coordinate with a human discussion partner.

However, challenges were also identified, such as occasional misinterpretation of user responses, difficulty handling tangential or off-topic remarks, and the need for more variety in discussion prompts to maintain long-term engagement. Some users also expressed a preference for human interaction and noted that the chatbot lacked the warmth and empathy of a real conversation partner.

Based on this initial feedback, the research team is iterating on the chatbot design to improve its natural language understanding, response generation, and prompt diversity. The project is open-sourced (https://github.com/Twilight-Tales/Twilight-Chat) and welcomes community contributions on advanced features such as adjustable speech rate, volume, and tone to accommodate users with hearing or cognitive impairments.

The next phase of the study will involve deploying the refined chatbot to the full intervention group and monitoring their use and cognitive outcomes over a one-year period. Participants will complete regular assessments of cognitive function, such as the Cognitive Ecological Momentary Assessment (EMA) \cite{singh2023ecological}, as well as a survey of their reading habits, engagement with the chatbot, and satisfaction with the experience at the end of the study. The control group will complete the same assessments without access to the chatbot, allowing for comparison of cognitive trajectories between the two groups.

However, the study has several limitations and challenges to consider. The small sample size and specific population (assisted living residents) may limit the generalizability of the findings to other aging adults. The long-term effectiveness and user retention for the chatbot remain to be seen, as novelty effects may wear off over time. There are also ethical considerations around data privacy, consent, and the potential for users to form attachments to or dependence on the AI system.

Continued research should aim to replicate the findings in larger and more diverse samples, explore the integration of objective measures of engagement and cognitive performance, and investigate the optimal design features and prompting strategies for maintaining long-term user benefit and adherence. As LLM chatbots become more sophisticated and human-like, ongoing work must also grapple with the ethical and social implications of their use in sensitive contexts such as elder care and mental health support.

\subsection{Case Study 3: Evaluating Open-Source LLMs for Medical Conversation Summarization}
This study\cite{chen2024comparative} focuses on evaluating the performance of two open-source LLMs, Llama2-70B \cite{touvron2023llama2} and Mistral-7B \cite{jiang2023mistral}, in summarizing three types of medical conversations, and benchmarked on corresponding datasets: MEDIQA-QS/MeQSum (consumer health questions), MEDIQA-ANS/MEDIQA-MAS (medical question-answering discussions), and iCliniq (doctor-patient dialogues). The aim is to assess the feasibility and quality of using open-source LLMs for extractive and abstractive summarization of medical conversations, which has applications in clinical decision support, patient education, and medical record summarization.

To evaluate the quality of the generated summaries, the study employs a novel pairwise comparison approach using GPT-4 \cite{openai2023gpt4} as an expert assessor model, and GPT-3.5 \cite{GPT3, instructGPT} as comparison target model. Such LLM-as-a-judge approaches have been reported in other literature, such as reference \cite{dhurandhar2024ranking}, but it remains a very new and active research area with many unknowns regarding its reliability and validity. Using a single target model as the benchmark reference is our unique innovation and contribution to this approach, aiming to lower the total number of pairwise comparisons and reduce the associated time and cost. For each test instance, the summaries generated by the two open-source LLMs are presented to GPT-4, which is prompted to compare them against the summary generated by GPT-3.5 on four key dimensions: relevance (selection of key information), coherence (logical flow and transitions), fluency (grammaticality and readability), and consistency (factual alignment with the input). GPT-4 selects the better summary for each pair and provides a brief rationale for its choice. This pairwise comparison is repeated with the order of the summaries swapped to control for positional bias.

Preliminary results on a subset of the test data suggest that Llama2 outperforms Mistral on most evaluation dimensions across the three datasets: Table \ref{tbl:summarization_result} presents the comparative performance of Llama2-70B and Mistral-7B against the GPT-3.5 baseline, with GPT-4 serving as the evaluator across five different medical datasets: MEDIQA-QS, MeQSum, MEDIQA-ANS, MEDIQA-MAS, and iCliniq. For each dataset, the table shows the win rates (as percentages) of Llama2-70B, Mistral-7B, and GPT-3.5, as well as the percentage of ties. 

\begin{table}[h]
\centering
\caption{Comparative performance of Llama2-70B and Mistral-7B versus GPT-3.5 across different datasets, evaluated by GPT-4.}
    \label{tbl:summarization_result}
    \centering
    \tabcolsep=0.12cm
\begin{tabular}{|c|c|c|c|c|c|c|}
\hline
\textbf{Dataset} & \multicolumn{3}{|c|}{\textbf{Llama2-70B Win Rate (\%)}} & \multicolumn{3}{|c|}{\textbf{Mistral-7B Win Rate (\%)}} \\
\cline{2-7}
                 & \textbf{Llama2} & \textbf{GPT-3.5} & \textbf{Tie} & \textbf{Mistral} & \textbf{GPT-3.5} & \textbf{Tie} \\
\hline
MEDIQA-QS        & 43\%                & 17\%             & 40\%         & 19\%                & 36\%             & 45\%         \\
\hline
MeQSum           & 42\%                & 18\%             & 40\%         & 14\%                & 51\%             & 35\%         \\
\hline
MEDIQA-ANS       & 43\%                & 22\%             & 35\%         & 40\%                & 24\%             & 36\%         \\
\hline
MEDIQA-MAS       & 40\%                & 38\%             & 22\%         & 31\%                & 38\%             & 31\%         \\
\hline
iCliniq          & 44\%                & 16\%             & 40\%         & 23\%                & 37\%             & 40\%         \\
\hline
\end{tabular}
\end{table}

The win rates indicate the proportion of instances where GPT-4 judged the summary generated by the respective model to be superior to the GPT-3.5 baseline. Ties represent cases where GPT-4 did not have a clear preference between the evaluated model and GPT-3.5. Across all datasets, Llama2-70B consistently outperforms GPT-3.5, with win rates ranging from 40\% to 44\%. In contrast, Mistral-7B only surpasses GPT-3.5 on the MEDIQA-ANS dataset with a 40\% win rate, but still falls short of Llama2-70B's 43\% win rate on the same dataset. For the other datasets, Mistral-7B's win rates range from 14\% to 31\%, consistently lower than both Llama2-70B and GPT-3.5. The percentage of ties is relatively high across all datasets and models, ranging from 22\% to 45\%. This suggests that in many cases, GPT-4 did not have a strong preference between the summaries generated by the evaluated models and the GPT-3.5 baseline.

Overall, the table demonstrates that Llama2-70B is the best-performing model among those evaluated, showing clear superiority over the GPT-3.5 baseline across all five medical datasets. Mistral-7B, while outperforming GPT-3.5 on MEDIQA-ANS, generally underperforms relative to both Llama2-70B and GPT-3.5 on the other datasets. Llama2's summaries were judged by GPT-4 to be more relevant and coherent, capturing the essential information from the input while maintaining a logical structure. Mistral's summaries, while generally fluent and consistent, tended to miss key details or introduce minor factual errors. Both LLMs struggled with highly technical or domain-specific language in the medical QA datasets, highlighting the need for further fine-tuning or incorporation of medical knowledge bases.

At the time of writing this manuscript, we have only completed the first round of computations and evaluations; thus, the numbers reported in the table are the result of a single experiment. However, we plan to conduct 3-5 repeats of the experiment to assess the mean and standard deviation of the results. This will help us ensure the robustness and reliability of our findings.

Qualitative analysis of GPT-4's pairwise judgments offers additional insights into the strengths and weaknesses of the LLMs for medical summarization. GPT-4 praised Llama2's summaries for their brevity, clarity, and ability to identify the most salient points from lengthy or complex input texts. However, it critiqued Llama2's tendency to occasionally overgeneralize or omit nuanced details. Mistral's summaries were noted for their natural, human-like language and maintenance of the input's overall tone and style, but were faulted for occasional redundancy or inclusion of tangential information.

This study demonstrates the potential of open-source LLMs for medical text summarization and highlights the value of using an expert LLM (GPT-4) for nuanced, qualitative evaluation of the quality of the generated summaries. The pairwise comparison approach allows for fine-grained assessment of specific language dimensions and provides interpretable rationales for model preferences, which can guide further improvement of the summarization models.

However, the study has several limitations. The evaluation relies on the judgment of a single LLM (GPT-4), whose own biases, knowledge gaps, or inconsistencies may affect the comparisons. At current stage, this study does not include human expert evaluation or comparison to ground-truth summaries, which would provide additional validation of the LLMs' performance. The datasets used, while diverse, may not represent the full range of medical text types or domains that summarization models would need to handle in practice.

Future work should incorporate multiple LLM evaluators and human expert judgments to robustly assess summary quality and validate the pairwise comparison approach. Expanding the evaluation to a wider range of medical text corpora and fine-tuning the LLMs on larger medical language datasets may improve their domain-specific performance. Exploring techniques for integrating medical knowledge into the summarization process (e.g., via knowledge graphs or retrieval-augmented generation) may help mitigate factual errors and inconsistencies.

As open-source LLMs continue to advance, their potential for medical text summarization grows. However, careful evaluation, domain adaptation, and integration with human expertise will be critical to ensure the reliability, safety, and utility of LLM-generated medical summaries in real-world clinical applications.

\subsection{Case Study 4: Designing an AI-Powered Patient Engagement System}
This case study presents the design and development of an AI-powered patient engagement system aimed at improving the management of chronic diseases through automated patient outreach. The system, currently being piloted for Crohn's Disease (CD) patients in collaboration with the Cleveland Clinic Foundation (CCF) and Morehouse School of Medicine (MSM), leverages large language models (LLMs) to generate and maintain longitudinal dialogues with patients via SMS or voice modalities.

The core of the system is a dialogue management engine powered by prompt-tuned LLMs that can understand patient responses, ask follow-up questions, provide guidance and education, and escalate to human health care personnel when needed.

\begin{figure}[htbp]
    \centering
    \includegraphics[width=1\linewidth]{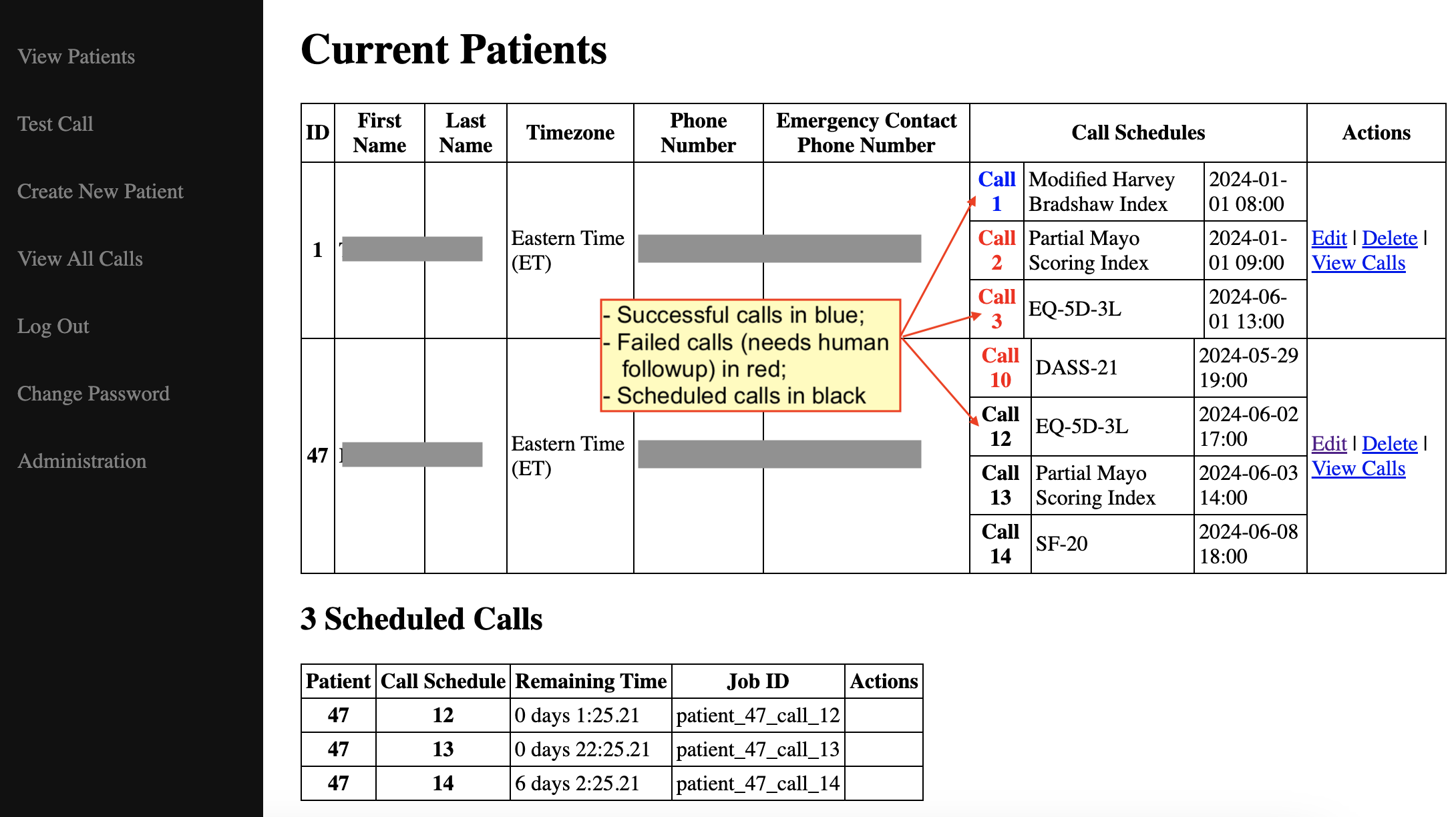}
    \caption{Clinician dashboard displaying current patients with their upcoming call schedules color-coded by status (completed, scheduled, failed), allowing easy monitoring of patient engagement.}
    \label{fig:4-patient-view}
\end{figure}

\begin{figure}[htbp]
    \centering
    \includegraphics[width=1\linewidth]{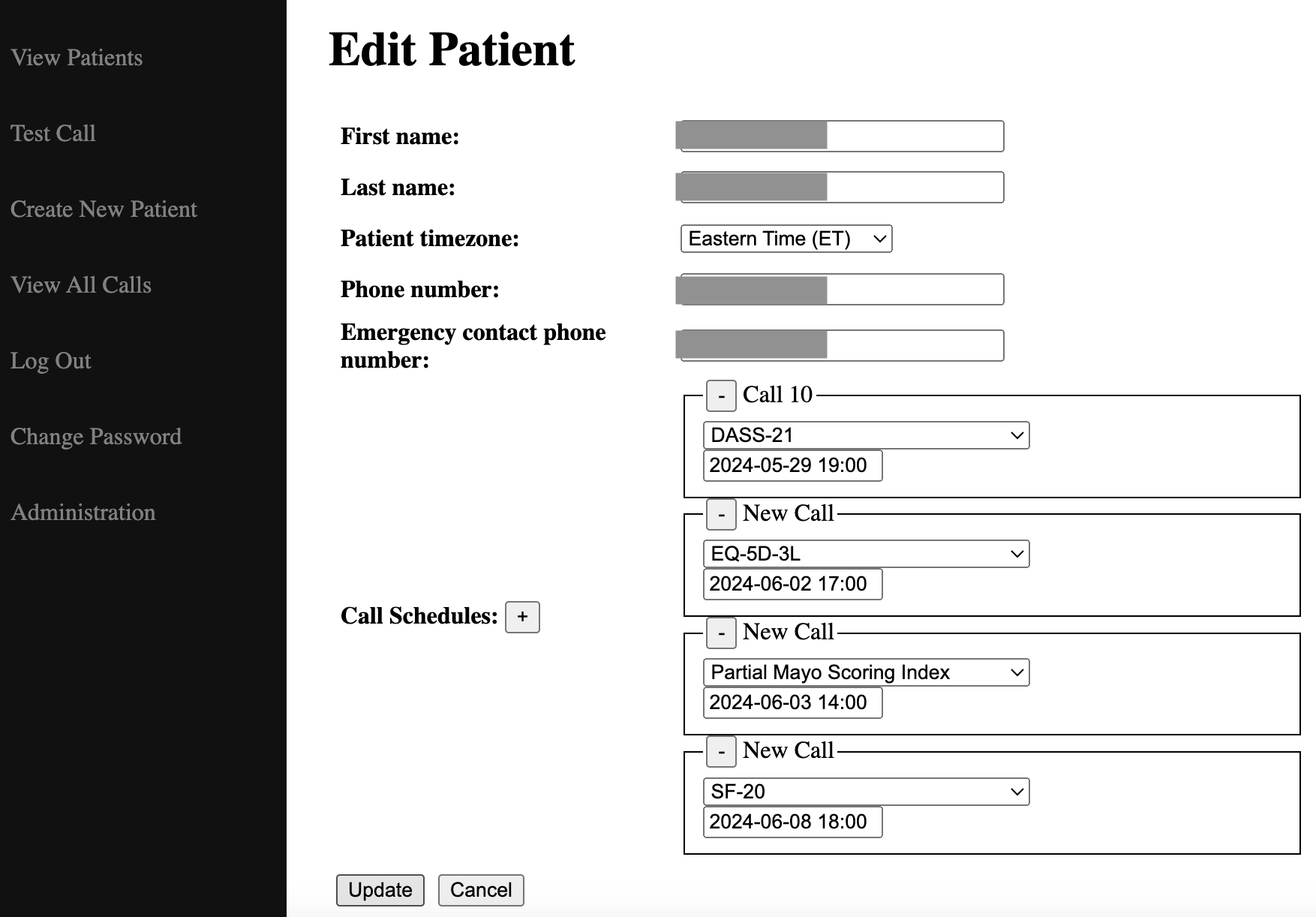}
    \caption{Patient profile editing screen enabling clinicians to update patient information, view upcoming scheduled calls, and add new calls with specific instruments like quality of life assessments.}
    \label{fig:4-edit-patient}
\end{figure}

Clinicians can monitor patient engagement through a web-based dashboard (Fig. \ref{fig:4-patient-view}) that displays current patients
along with their upcoming call schedules, color-coded to indicate status such as completed, scheduled, or failed
calls. Drilling down into an individual patient profile (Fig. \ref{fig:4-edit-patient}), clinicians can update patient information, view
upcoming scheduled calls, and add new calls specifying the desired instruments to be administered, such as disease-specific quality of life assessments.


\begin{figure}[htbp]
    \centering
    \includegraphics[width=1\linewidth]{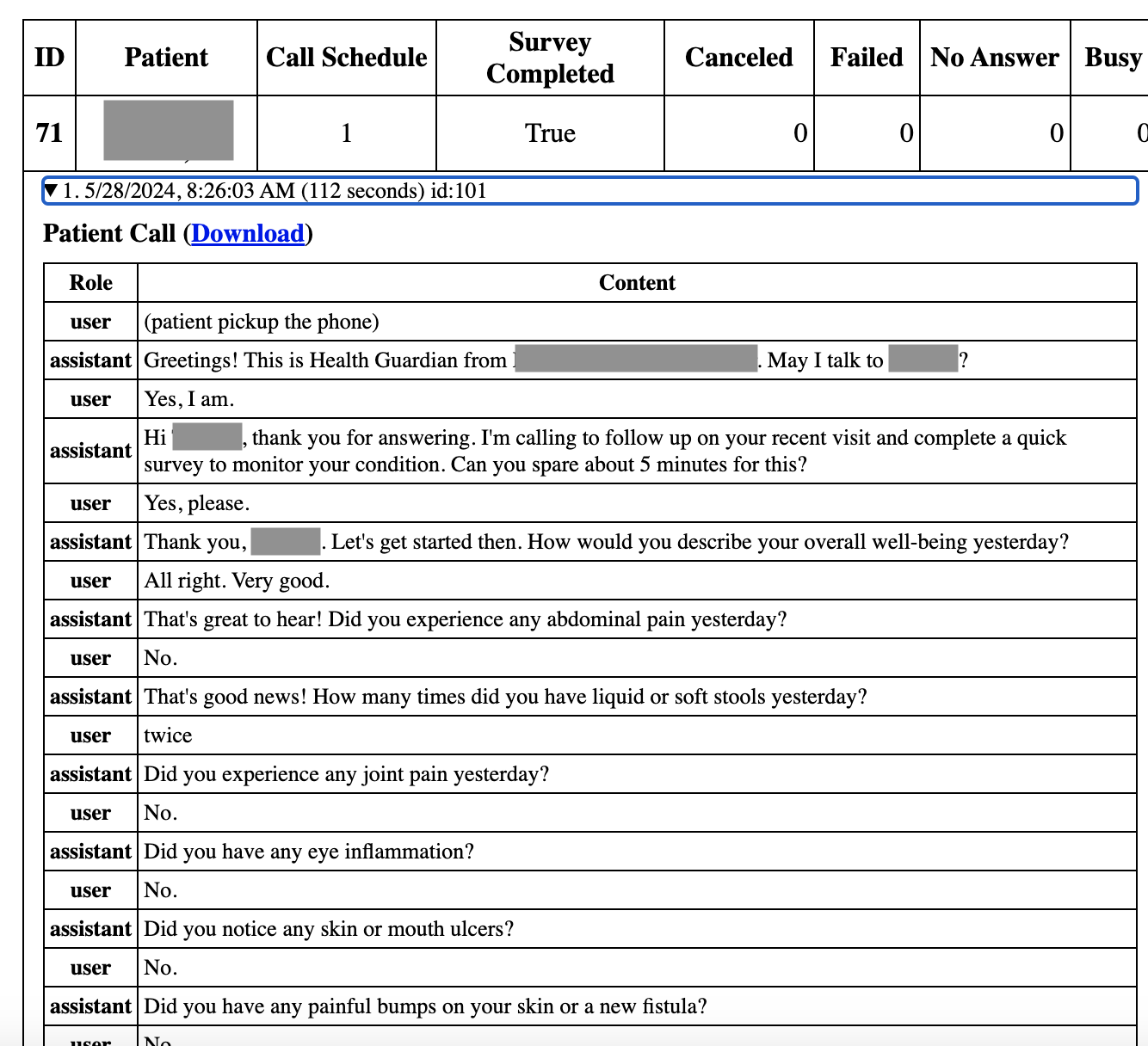}
    \caption{UI for reviewing of the patient chat history, showing a transcription of the chatbot conversing
    naturally with the patient to ask follow-up questions about their well-being, symptoms, and quality of life.}
    \label{fig:4-chat-history}
\end{figure}

At the time of scheduled call, the backend engine will use VoIP providers such as Twilio to create a call or SMS
session with the patient, then connect the session to the LLMs chatbot. The LLM-driven conversations aim to elicit
information about patients' symptoms and quality of life by conducting standard medical questionnaires in casual
conversations, aiming to encourage patients to share their health status and providing clinicians with a more
comprehensive and timely understanding of patient health status between visits. As shown in Fig. \ref{fig:4-chat-history}, the AI assistant
engages in natural conversation with patients, seamlessly administering standard assessments while also inquiring
about their general well-being, symptoms, and quality of life. Prompt engineering techniques guide the LLMs in
following specific call scripts to meet clinician-defined task assignments, such as conducting a particular
questionnaire or post-treatment follow-up. When the call is finished, the chat history will be stored into a database
and can be reviewed by the health team on the web portal (Figure \ref{fig:4-chat-history}).

\begin{figure}[htbp]
    \centering
    \includegraphics[width=1\linewidth]{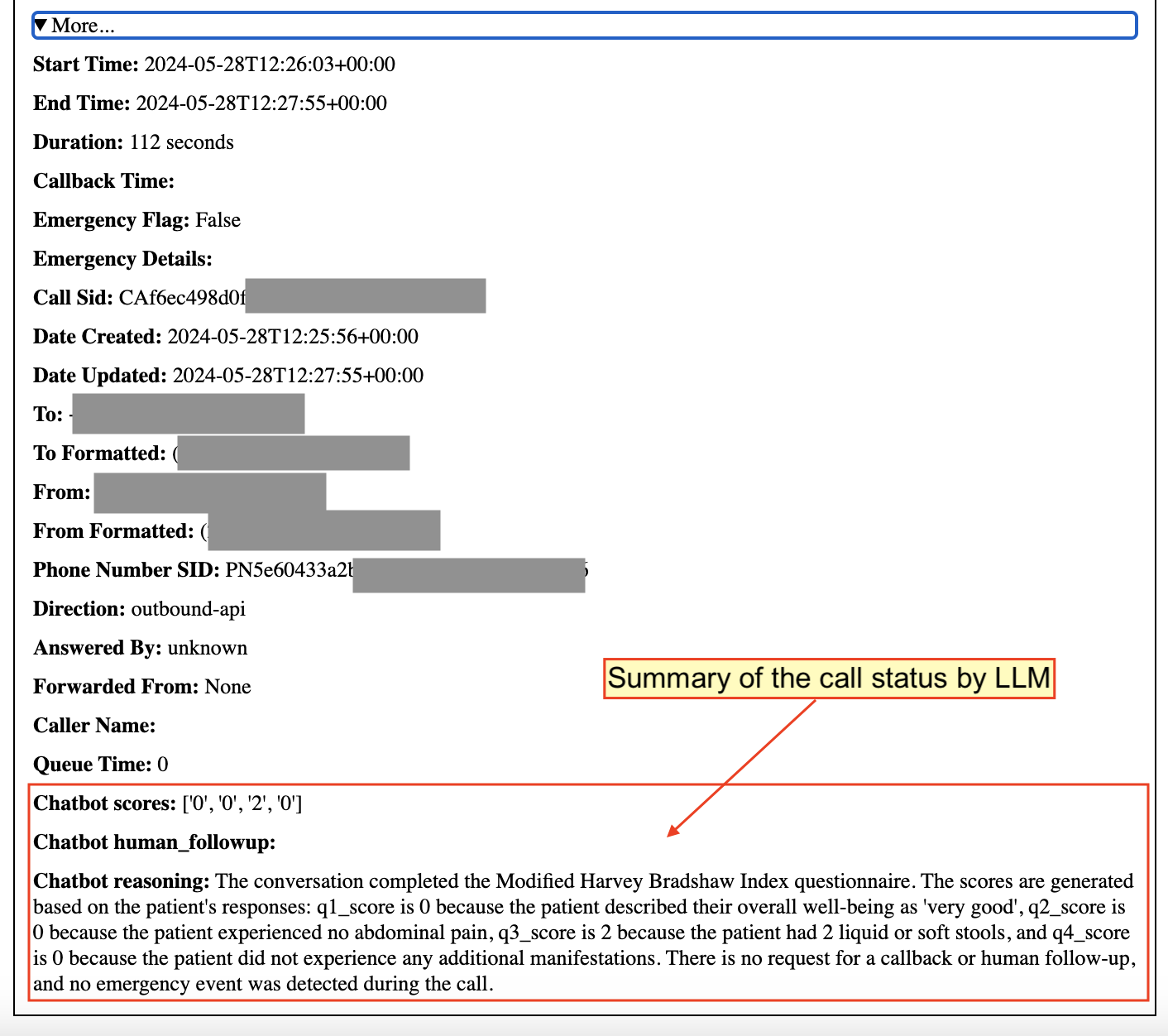}
    \caption{Call summary generated by the AI system, providing key details such as call duration, chatbot scores reflecting the patient's responses on standard assessments, chatbot reasoning explaining the scores, and noting no emergency events or callback requests.}
    \label{fig:4-chat-summary}
\end{figure}

The system will then automatically generate call summaries (Fig. \ref{fig:4-chat-summary}) that provide clinicians with key information
at a
glance, including call duration, chatbot scores reflecting patient responses on standard instruments, and explanations of the scores based on the patient's replies. Importantly, the system also flags any emergency events or patient requests for human callback, allowing clinicians to quickly identify and prioritize cases needing attention.

By presenting this functionality through the above illustrative screenshots of the actual system, we aim to provide
readers with a concrete understanding of how LLM-powered patient engagement can be realized in practice. The images demonstrate the system's ability to support natural, contextualized patient dialogue, automated reporting of patient state, and seamless integration with clinical workflows through an intuitive interface.

The system can be further improved by fine-tuning the LLMs on large datasets of medical conversations and augmenting it with domain-specific knowledge bases to ensure the accuracy and relevance of their responses. By implementing retrieval-augmented generation technique, it is possible to allow the system to personalize the dialogue style and content to each patient's individual characteristics, preferences, and communication history. The system can maintain a dynamic patient profile that includes demographic information (e.g., age, preferred language, health literacy), clinical data (e.g., diagnoses, medications, lab reports), and interaction data (e.g., tone, sentiment, topics discussed). By dynamically loading the patient profile into each conversation's system prompt, the system can control the LLMs' language use to tailor the conversation to the patient's needs and context.

As this system progresses from alpha testing through clinical beta testing and initial patient deployment, we will
continue to gather data on its usability, performance, and impact on patient engagement and outcomes. We look forward
to sharing further insights and results from this ongoing work at ICDH 2025.

\section{Discussion and Future Directions}
The case studies presented in this paper demonstrate the significant potential of LLMs in building digital health solutions to transform various aspects of healthcare, from patient engagement and clinical decision support to medical knowledge synthesis and mental health assessment. By enabling more natural, efficient, and scalable interactions between patients, providers, and medical information, LLMs are poised to enhance the accessibility, personalization, and effectiveness of digital health interventions\cite{johnsnowlabs2023power, topflightapps2023llms}.

However, realizing the full potential of LLMs in healthcare requires addressing many more challenges and considerations that are not covered or only slightly explored by the 4 use cases: Ensuring the privacy and security of sensitive patient data used to train and apply LLMs is critical, and techniques like data de-identification, federated learning, and secure pipelines must be employed\cite{johnsnowlabs2023power, lantanablog2023llms}. Mitigating bias and promoting fairness in LLM outputs is another essential priority, necessitating diverse training datasets, regular auditing, and inclusive development practices\cite{ueda2024fairness}.

The ``black box'' nature of LLMs also raises concerns around transparency and interpretability, particularly for high-stakes medical decisions. Explainable AI techniques, clear communication of limitations, and human oversight are key to building trust and accountability in LLM-assisted healthcare\cite{gaper2023interpretable, frontiers2022transparency}. Regulatory frameworks and evaluation standards specifically tailored to the unique considerations of medical LLMs are also needed to ensure their safe and responsible deployment\cite{nature2023navigate, himss2023regulatory}. LLM-as-a-judge, which discussed in use case 3, or LLM-as-a-explainer, are potential solutions for these concerns.

Looking ahead, there are numerous promising opportunities for LLMs to transform digital health. Emerging applications include virtual mental health support, real-time clinical decision assistance, automated medical data analysis, and AI-powered patient engagement platforms\cite{nixon2024vision, johnsnowlabs2023power, topflightapps2023llms}, like we discussed in use case 1,2 and 4. LLMs could be combined with other cutting-edge technologies like Internet of Medical Things devices and blockchain systems to enable even more intelligent and secure healthcare solutions\cite{magnimind2023aiblockchain, waheed2023fedblockhealth, 9906419}.

Personalizing LLMs to individual patients by integrating multimodal data sources such as EHRs, EMRs, sensor streams, and genomic profiles is another exciting frontier. Such tailored LLMs could provide hyper-personalized health insights, recommendations, and interventions, potentially revolutionizing precision medicine and chronic disease management. Fine-tuning LLMs on specialized medical knowledge bases and augmenting them with domain-specific reasoning capabilities could also significantly enhance their utility for complex clinical decision support and research applications\cite{nixon2024vision, pmc2023evalmedLLM, diagnosticpathology2024critique}.

However, unlocking the next generation of medical LLMs will require concerted collaborations between AI researchers, healthcare professionals, policymakers, and patient advocates. Developing standardized benchmarks and evaluation frameworks for assessing the performance, safety, and ethics of medical LLMs across diverse healthcare settings and populations is crucial\cite{nature2023navigate}. Open-sourcing model architectures, training pipelines, and validation datasets would democratize access and accelerate innovation in this space\cite{lantanablog2023llms}.

Interdisciplinary teams must also work together to proactively surface and address the societal implications of medical LLMs, from data governance and consent processes to liability frameworks and health equity considerations\cite{pmc2023digital}. Cultivating a digitally literate healthcare workforce and educating patients about the benefits and limitations of LLM-assisted care will also be key to successful adoption and impact\cite{pmc2023digital}.

By bringing together the medical, technical, and ethical expertise needed to responsibly develop and deploy LLMs, the digital health community can pioneer a new paradigm of AI-enabled healthcare that empowers patients, augments providers, and optimizes health outcomes for all.

\section{Conclusion}
The case studies and applications presented in this paper demonstrate the significant potential of large language models to transform various aspects of digital health, from patient engagement and clinical decision support to medical knowledge synthesis and mental health assessment. By enabling more natural, efficient, and intelligent interactions between patients, providers, and medical information, LLMs have the power to enhance the accessibility, personalization, and effectiveness of healthcare delivery.

However, realizing the full promise of LLMs in healthcare requires addressing multifaceted challenges related to data privacy, bias, transparency, and regulatory compliance. Ensuring the responsible development and deployment of medical LLMs will demand ongoing interdisciplinary collaboration among healthcare stakeholders, AI researchers, policymakers, and patient advocates.

Key priorities include establishing robust data governance frameworks, model auditing protocols, and human-AI interaction guidelines tailored to healthcare contexts. Developing standardized evaluation methodologies and benchmarks for assessing the performance, safety, and fairness of medical LLMs across diverse patient populations and clinical settings is also critical. Additionally, investing in AI literacy training for healthcare professionals and public education about the benefits and risks of LLM-assisted care will be essential for driving adoption and impact.

Looking ahead, the next frontiers for medical LLMs lie in advancing personalized, specialized, and multimodal applications that can transform the diagnosis, treatment, and prevention of complex and costly health conditions. Achieving these ambitious goals will require sustained research and development efforts, guided by a shared commitment to ethical principles, rigorous evidence, and equitable impact.

By bringing together the unique strengths and perspectives of diverse stakeholders across the healthcare ecosystem, we can responsibly harness the power of language AI to improve patient outcomes, enhance provider experiences, and optimize health systems performance at global scale. As the field of digital health continues to evolve, it is imperative that we proactively shape the development and deployment of LLMs to ensure that their transformative potential benefits all patients and communities.

\section*{Acknowledgments}
\subsection*{AI-Generated Content Disclosure}
This paper was prepared with the assistance of artificial intelligence (AI) tools. Perplexity.ai, a literature search and discovery platform, was utilized to conduct a comprehensive literature review and gather relevant research papers and articles. Claude, an AI assistant developed by Anthropic, was employed for drafting the initial version of the paper. GPT-4, an AI model created by OpenAI, provided reviewing and improvement suggestions.

While AI assistance was instrumental in the preparation of this paper, all content was carefully reviewed, edited, and approved by the human authors, who take full responsibility for the final manuscript.

\subsection*{Collaborator Acknowledgments}
We  express our sincere gratitude to our collaborators, Tesfaye Yadete from CCF, Kingsley Njoku from MSM, Yuhao Chen and Zhimu Wang from Queen's University Canada, Hannah Zhou from Cornell University for their valuable contributions to the research. Bo Wen also extend sincere appreciation to Dr. Jeffrey L. Rogers, for his unwavering support and encouragement in pursuing this work.

\bibliographystyle{IEEEtran}
\bibliography{main}

\end{document}